\documentclass{article}

\usepackage{arxiv}

\usepackage[utf8]{inputenc} 
\usepackage[T1]{fontenc}    
\usepackage{hyperref}       
\usepackage{url}            
\usepackage{booktabs}       
\usepackage{amsfonts}       
\usepackage{nicefrac}       
\usepackage{microtype}      
\usepackage{lipsum}		
\usepackage{graphicx}
\usepackage{natbib}
\usepackage{doi}
\usepackage{color}
\usepackage{amsmath}
\usepackage{mathtools}

\DeclareMathOperator{\Tr}{Tr}

\title{Bayesian Information Criterion for Event-based Multi-trial Ensemble Data}

\author{ \href{https://orcid.org/0000-0002-3027-0090}{\includegraphics[scale=0.06]{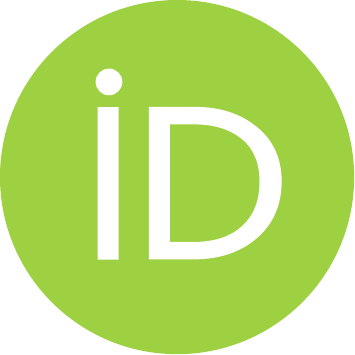}\hspace{1mm}Kaidi Shao}
	\\
	Department of Physiology of Cognitive Processes\\
	Max-Planck Institute for Biological Cybernetics\\
	Tuebingen, 72076, Germany \\
	International Center for Primate Brain Research\\
	Songjiang, Shanghai, 201602, China\\
	\texttt{kaidi.shao@tuebingen.mpg.de} \\
	\And
	{\hspace{1mm}Nikos K. Logothetis}\\
	International Center for Primate Brain Research\\
	Songjiang, Shanghai, 201602, China\\
	Department of Physiology of Cognitive Processes\\
	Max-Planck Institute for Biological Cybernetics\\
	Tuebingen, 72076, Germany \\
	Centre for Imaging Sciences, Biomedical Imaging Institute\\
	The University of Manchester\\
	 Manchester, UK\\
	\texttt{nikos.logothetis@tuebingen.mpg.de} \\
	\And
	\href{https://orcid.org/0000-0003-0025-2323}{\includegraphics[scale=0.06]{orcid.pdf}\hspace{1mm}Michel Besserve} \\
	Department of Empirical Inference\\
	Max-Planck Institute for Intelligent Systems\\
	Tuebingen, 72076, Germany \\
	\texttt{michel.besserve@tuebingen.mpg.de} \\
}

\renewcommand{\undertitle}{}
\renewcommand{\shorttitle}{Multi-trial BIC}

\hypersetup{
pdftitle={A template for the arxiv style},
pdfsubject={q-bio.NC, q-bio.QM},
pdfauthor={David S.~Hippocampus, Elias D.~Striatum},
pdfkeywords={First keyword, Second keyword, More},
}

\begin{document}
\maketitle

\begin{abstract}
Transient recurring phenomena are ubiquitous in many scientific fields like neuroscience and meterology. Time inhomogenous Vector Autoregressive Models (VAR) may be used to characterize peri-event system dynamics associated to such phenomena, and can be learned by exploiting multidimensional data gathering samples of evolution of the system in multiple time windows comprising, each associated to one occurrence of the transient phenomenon, that we will call ``trial''. However, optimal VAR model order selection methods, commonly relying on the Akaike or Bayesian Information Criteria (AIC/BIC), are typically not designed for multi-trial data. Here we derive the BIC methods for multi-trial ensemble data which are gathered after detection of the events. We show using simulated bivariate AR models that the multi-trial BIC is able to recover the real model order. We also demonstrate with simulated transient events and real data that the multi-trial BIC is able to estimate a sufficiently small model order for dynamic system modelling.
\end{abstract}


\section{Introduction}

Transient dynamical events, such as weather extremes in climate science \citep{grotjahn2016north} or reactivations of hippocampal memory traces in the mammalian brain \citep{sutherland2000memory}, are ubiquitous phenomena in non-linear dynamical systems.  
Typically, they are prone to occur repetitively during specific dynamic regimes, and may contribute to critical system level modifications. Understanding their generation mechanisms and the effects they trigger requires precise modelling of the transient dynamics based on transient event data, i.e., the multi-trial ensemble datasets gathered after detecting these events in the observation of the systems. 

An effective way to model transient dynamics is to use time-inhomogeneous Vector Autoregressive (VAR) which are able to approximate the non-linear dynamics in a particular region of the state space with a time-varying linear model. After gathering a multi-trial dataset, composed of samples of the system dynamics during multiple occurrences (named trials) of a type of event, the VAR model can be determined by estimating the model parameters with the well-established Ordinary Least Square (OLS) or Recursive Least Square methods \citep{hesse_2003, mller_2001}.   

However, the design of appropriate VAR model order selection criterion for such multi-trial ensemble data remains largely unaddressed. 
Accurate model order estimation is critical as: 
1) a too low order will prevent gathering sufficient past dynamics to predict future values and lead to larger residual covariance estimates. 
2) a too large order may lead to over-fitting and spurious parameter values which may affect downstream tasks (forecasting, causality analysis, ...).

Two common ways to optimize model order to model single trial signals are the Akaike information criterion (AIC) \citep{Akaike1974ANL,akaike1998information} and the Bayesian information criterion (BIC) \citep{gideon1978estimating}. 
They both introduce a penalty term in the $\log$-likelihood function to compensate for the effect caused by over-fitting with over-complex models. 
The penalty term is dependent on the number of parameters and the sample size, and determining these two quantities causes difficulties for the multi-trial case compared to the single trial case. 

Previous studies proposed various methods to estimate time-varying VAR model (for an overall review see \citep{cekic_2018a}), together with some model order selection criteria for each specific type of methods. However, they did not provide a specific penalty term which is applicable for non-stationary and multi-trial cases.
For example, Ding et al. proposed the short windowing method which assumes local stationarity in the process \citep{ding_2000}, which is also a ML estimator of the coefficients in a time window. 
Model order in this study was optimized by the classical AIC by assuming total likelihood $\mathcal{L}(p)$ is the product of the likelihood of each sample in all the trials. However, the penalty term was not revised accordingly. 
In a study using adaptive methods \citep{hesse_2003}, a similar treatment was applied. 
Another related paper suggested to perform model order selection individually for each short window, and select the maximum order for the whole process \citep{mller_2001}. 
While this idea might sound reasonable, this likely leads to a very large model order in settings with large number of short windows. This is especially the case with time-inhomogeneous models, where coefficients are changing at each time point (i.e. window length is 1 time point). 
\citet{rajan1996generalized} and \citet{eom_1999} derived in a Bayesian approach the criteria for time-varying VAR models, which is similar to our proposal below. However, they assume that the coefficient matrix is composed of some basis functions, and the criteria depend on the number of basis functions, which differs from the classical BIC form. 

In this note, we provide an extended version of BIC appropriate for non-stationary signals with multi-trial structure. The BIC is derived by assuming a vector autoregressive model of the data.
We demonstrate the effectiveness of optimal model order selection with simulated and empirical data. 


\section{Methods}

In the Methods section, we first define what is meant by multi-trial ensemble data and the associated VAR model. Then the Maximum Likelihood Estimation of the model parameters are provided, together with the Hessian matrix of the coefficient matrix, as a prerequisite for further derivation of multi-trial BIC in Section~\ref{proof: multi-trial BIC}. 

\subsection{Vector Autoregressive Modelling of Multi-trial ensemble data} \label{sec: multi-trial VAR}

For a given system (physical, biological, ...), we assume we have a long time series of $d$-dimensional observations denoted $\tilde{\boldsymbol{X}}_t$.  
Classically, specific patterns in the observations, referred to as events, are located in $\tilde{\boldsymbol{X}}_t$ by a detection procedure, yielding a list of time points at which the events occur, called \textit{reference points}.  
For each \textit{reference point} $t_n$, 
the samples from the long time series $\tilde{\boldsymbol{X}}_t$ covering a fixed peri-event time window $t_n+\mathcal{I}$ of length $T$ around $t_n$, with e.g. $\mathcal{I}=[-T/2,\,T/2-1]$ ($T$ assumed even for simplicity), are extracted to build a two-way panel $\{\boldsymbol{X}^{(n)}_{t^\prime}\}$:
\begin{equation} \label{eq:periEventPanel}
\boldsymbol{X}_{t^\prime}^{(n)} = \tilde{\boldsymbol{X}}_{t^\prime+t_n}, t^\prime \in \mathcal{I}, t_n\in \mathcal{T}\,.
\end{equation}
This panel forms the multi-trial ensemble dataset with length $T$ and trial number $N$. 
In the following, we denote this ensemble dataset as $\{\boldsymbol{X}_{t}^{(n)}\}_{t=1...T}$ and the samples for all trials at time $t$ as  $\{\boldsymbol{X}_{t}^{(n)}\}$.
These extracted trials can be understood as repetitive realizations of the same stochastic processs, whose time-varying variable is referred to as $\boldsymbol{X}_t$.

We then assume the transient, possibly non-linear, dynamics underlying  $\{\boldsymbol{X}_{t}^{(n)}\}_{t=1...T}$ can be approximated for all $n$ by a common time-inhomogeneous Vector Autoregressive (VAR) model: 
\begin{equation}\label{eq:varinhomo3}
\boldsymbol{X}_t = A_t\boldsymbol{X}_{p,t}+\boldsymbol{\eta}_t\,,\boldsymbol{\eta}_t\sim \mathcal{N}(\boldsymbol{k}_t,\Sigma_{t})\,
\end{equation}
where $\boldsymbol{X}_{p,t}=\left[\boldsymbol{X}_{t-1}^\top,\cdots,\boldsymbol{X}_{t-p}^\top\right]^\top$ collects past process values up to lag $p$ in a $pd$-dimensional vector, while $p$ is referred to as model order, and $A_t$ is a time dependent matrix of appropriate dimension and
$\{\boldsymbol{\eta}_t\}_{t\in \mathbb{Z}}$ is called the \textit{innovation} process.  
Innovations at each time points are assumed to be jointly independent $n$-dimensional Gaussian random vectors. Moreover, the covariance $\Sigma_t$ between the components is assumed diagonal.

The model is determined by the model order $p$ and 3 model parameters: the coefficient matrix $A_t$, the innovations mean $\boldsymbol{k}_t$ and the innovations covariance $\Sigma_{t}$.
These model parameters are assumed time-inhomogenous, notably allowing to approximate the non-linearity of the evolution of the system within the event time-window.
Importantly, the VAR model entails assumptions allowing the estimation of the parameters from data. Chiefly, the independence assumption between innovations at different time points entails order-$p$ Markovianity of the process, as the distribution of $\boldsymbol{X}_t$ given the whole history of the process up to time $t-1$ depends only on $\boldsymbol{X}_{p,t}$.

\subsection{Estimation of the model from data}

\subsubsection{Maximum Likelihood Estimation of VAR parameters} \label{sec:var_estim}

In this section we provide a Maximum Likelihood (ML) estimation of the model parameters for VAR(p) model described in Section~\ref{sec: multi-trial VAR} for the multi-trial ensemble data.

For a dataset with $N$ \textit{i.i.d} trials, the likelihood function is the product of the likelihood of each sample across time points and trials
\[
\mathcal{L}\left(A_{t}, \boldsymbol{k}_t, \Sigma_{t} ; \{\boldsymbol{X}_{t}^{(n)}\}_{t=1...T}\right) = \prod_{t=1}^{T}\prod_{n=1}^{N} \frac{1}{\sqrt{ (2 \pi)^d \left |\Sigma_{t}\right |}} \exp \left(\!-\frac{1}{2}\left(\boldsymbol{X}_{t}^{(n)}\!-\!A_{t} \boldsymbol{X}_{p,t}^{(n)}-\boldsymbol{k}_t\right)^{\top} \Sigma_{t}^{-1}\!
\left(\boldsymbol{X}_{t}^{(n)}\!-\!A_{t} \boldsymbol{X}_{p, t}^{(n)}-\boldsymbol{k}_t\right)\!\right)\,.
\]

As $\boldsymbol{k}_t = \mathbb{E}[\boldsymbol{X}_t] - A_{t}\mathbb{E}[\boldsymbol{X}_{p,t}]$, the quadratic term can be approximately rewritten as a function of the demeaned samples $\overline{\boldsymbol{X}}^{(n)}_t =  \boldsymbol{X}_{t}^{(n)} - \widehat{\mathbb{E}}[\boldsymbol{X}_t]$, $\overline{\boldsymbol{X}}^{(n)}_{p,t} =  \boldsymbol{X}_{p,t}^{(n)} - \widehat{\mathbb{E}}[\boldsymbol{X}_{p,t}]$ using the empirical averages  $\widehat{\mathbb{E}}[\boldsymbol{X}_t]=\frac{1}{N}\sum^N_{n=1} {\boldsymbol{X}^{(n)}_t}$ and
$\widehat{\mathbb{E}}[\boldsymbol{X}_{p,t}]=\frac{1}{N}\sum^N_{n=1} {\boldsymbol{X}^{(n)}_{p,t}}$.
 The corresponding likelihood function is
\[
\mathcal{L}\left(A_{t}, \Sigma_{t} ; \{\boldsymbol{X}_{t}^{(n)}\}_{t=1...T}\right) = \prod_{t=1}^{T}\prod_{n=1}^{N} 
\frac{1}{\sqrt{(2 \pi)^d \left |\Sigma_{t}\right |}} 
\exp \left(-\frac{1}{2}\left(\overline{\boldsymbol{X}}_{t}^{(n)}-A_{t} \overline{\boldsymbol{X}}_{p,t}^{(n)}\right)^{\top} \Sigma_{t}^{-1}\right.\\
\left.\left(\overline{\boldsymbol{X}}_{t}^{(n)}-A_{t} \overline{\boldsymbol{X}}_{p, t}^{(n)}\right)\right)\,,
\]

The log likelihood thus takes the form
\[
\begin{aligned}
l\left(A_{t},  \Sigma_{t} ;  \{\boldsymbol{X}_{t}^{(n)}\}_{t=1...T}\right)
=& \log \mathcal{L}\left(A_{t}; \Sigma_{t} ;  \{\boldsymbol{X}_{t}^{(n)}\}_{t=1...T}\right) \\
=&-\frac{NTd}{2} \log (2 \pi)-\frac{NT}{2} \log \left|\Sigma_{t}\right |-\frac{1}{2} 
\sum_{t=1}^{T}\sum_{n=1}^{N}\left(\overline{\boldsymbol{X}}_{t}^{(n)}-A_{t} \overline{\boldsymbol{X}}_{p, t}^{(n)}\right)\\ &\Sigma_t^{-1} \left(\overline{\boldsymbol{X}}_{t}^{(n)}-A_{t} \overline{\boldsymbol{X}}_{p, t}^{(n)}\right)^{\top}\,.
\end{aligned}
\]
The ML estimation of the AR coefficient matrix $A_t$ at each time point $t$ requires finding the parameter values such that the following first-order derivative vanishes
\begin{equation}\label{eq:diff_logL_At}
\frac{d l\left(A_{t}, \Sigma_{t} ; \{\boldsymbol{X}_{t}^{(n)}\}_{t=1...T}\right)}{d A_{t}}=-2 \sum_{n=1}^{N} \Sigma_{t}^{-1}\left(\overline{\boldsymbol{X}}_{t}^{(n)}-A_{t}
\overline{\boldsymbol{X}}_{p,t}^{(n)}\right) {\overline{\boldsymbol{X}}_{p, t}^{(n)}}^{\top}=0\,.
\end{equation}

Reorganizing the terms yields
\[
\sum_{n=1}^{N}\left(\overline{\boldsymbol{X}}_{t}^{(n)}-\widehat{A}_t\overline{\boldsymbol{X}}_{p,t}^{(n)}\right) {\overline{\boldsymbol{X}}_{p, t}^{(n)}}^{\top}=0 \,.
\]
Thus the coefficient matrix can be estimated in the following form
\[
\widehat{A_t}=\left(\sum_{n=1}^{N} \overline{\boldsymbol{X}}_{t}^{(n)} {\overline{\boldsymbol{X}}_{p,t}^{(n) }}^{\top}\right)\left(\sum_{n=1}^{N} \overline{\boldsymbol{X}}_{p, t}^{(n)} { \overline{\boldsymbol{X}}_{p, t}^{(n)}}^{\top}\right)^{-1}.
\]
This term is equivalent to 

\begin{equation} \label{eq:coeff_estim}
\widehat{A_t} = \widehat{\Sigma}_{\boldsymbol{X}_t \boldsymbol{X}_{p}}( \widehat{\Sigma}_{\boldsymbol{X}_p})^{-1}\,,
\end{equation}
where the covariance matrices are estimated from the $N$-sampled data as: 
\begin{equation}
\widehat{\Sigma}_{\boldsymbol{X}_t \boldsymbol{X}_p}=\frac{1}{N}\sum^N_{n=1} {(\boldsymbol{X}^{(n)}_t-\widehat{\mathbb{E}}[\boldsymbol{X}_t]) (\boldsymbol{X}^{(n)}_{p,t}-\widehat{\mathbb{E}}[\boldsymbol{X}_{p,t}])^{\top} }\,,
\end{equation}
and
\begin{equation} \label{eq:Sigma_Xp}
\widehat{\Sigma}_{\boldsymbol{X}_{p}}=\frac{1}{N}\sum^N_{n=1} {(\boldsymbol{X}^{(n)}_{p,t}-\widehat{\mathbb{E}}[\boldsymbol{X}_{p,t}]) (\boldsymbol{X}^{(n)}_{p,t}-\widehat{\mathbb{E}}[\boldsymbol{X}_{p,t}])^{\top} }\,.
\end{equation}

The innovations' mean and covariance, estimated using the residual mean and residual covariance matrix, thus take the following form:
\begin{equation} \label{eq:inno_mean}
\widehat{\boldsymbol{k}_t} =\widehat{\mathbb{E}}[\boldsymbol{X}_t]-\widehat{A}_t\widehat{\mathbb{E}}[\boldsymbol{X}_{p,t}]\,,
\end{equation}
\begin{equation} \label{eq:inno_cov}
\widehat{\Sigma_t} =\frac{1}{N} \sum^{N}_{n=1} {(\boldsymbol{X}^{(n)}_t-\widehat{A_t}\boldsymbol{X}^{(n)}_{p,t}-\widehat{\boldsymbol{k}_t})  (\boldsymbol{X}^{(n)}_t-\widehat{A_t}\boldsymbol{X}^{(n)}_{p,t}-\widehat{\boldsymbol{k}_t})^{\top}}\,.
\end{equation}

\subsubsection{Hessian of the likelihood function for the coefficient matrix $A_t$} \label{proof: hessian}

The Hessian matrix can be obtained by calculating the derivative of the first-order derivative of the $\log$-likelihood given in Eq.~\ref{eq:diff_logL_At}, which can be rewritten as:
\begin{equation} \label{eq:1-order_deriv_logL_1trial_time-invariant_rewritten}
\frac{d \mathit{l}}{d A_t}
= \sum_{n=1}^{N} \Sigma_{t}^{-1} \overline{\boldsymbol{X}}_{t}^{(n)}  {\overline{\boldsymbol{X}}_{p, t}^{(n)}}^{\top}
-A_{t} {\overline{\boldsymbol{X}}_{p, t}^{(n)}} {\overline{\boldsymbol{X}}_{p, t}^{(n)}}^{\top}\,.
\end{equation}

We define a time-varying quantity $C_t$ as the sum of the covariance matrix of the lagged state $\boldsymbol{X}_{p,t}$:
\begin{equation}
C_t = \sum_{n=1}^{N}  {\overline{\boldsymbol{X}}_{p, t}^{(n)}} {\overline{\boldsymbol{X}}_{p, t}^{(n)}}^{\top} \,.
\end{equation}
Notably, if N is very large and asymptotically, the covariance of $p$ lagged-states for each time point $t$, supposed to be a constant matrix  with the dimension of $pd$, can be estimated by $ \widehat{\Sigma}_{\boldsymbol{X}_{p}} = \frac{C_t}{N} $, thus
\begin{equation} \label{eq:assymp}
C_t =  N \widehat{\Sigma}_{\boldsymbol{X}_{p}}\,.
\end{equation}

Each entry of the matrices $A_t$, $\Sigma_t^{-1}$ and $C_t$ at the $\mathit{i}^{th}$ row and $\mathit{j}^{th}$ column can be denoted as $a_{ij}$, $\sigma_{ij}$, $c_ij$. Then the second-order derivative of the $\log$-likelihood w.r.t. element $a_{ij}$ and $a_{km}$ is:
\begin{equation}
\frac{d^2 \mathit{l}}{d a_{ij} d a_{km}} = - c_{mi} \cdot \sigma_{kj}\,.
\end{equation}

The $d \times pd$ dimensional coefficient matrix $A_t$ can be vectorized as a $pd^2$-dimensional vector by concatenating each column. If the Hessian matrix is organized as $H=\frac{d^2 \mathit{l}}{d \mathit{vec}(A_t)^\top d \mathit{vec}(A_t)}$, then the Hessian matrix can be rewritten compactly as a Kronecker product:
\begin{equation} \label{eq:hessian_compact_homo}
H_t = - C_t \otimes \Sigma_t^{-1}\,.
\end{equation}

\subsection{BIC for multi-trial VAR models} \label{proof: multi-trial BIC}

\subsubsection{General form for BIC}
The remaining aspect to determine an optimized VAR model for the ensemble dataset is the model order. 
As briefly mentioned in the Introduction, classically proposed information criteria like AIC or BIC both depend on a penalized log likelihood function with the ML estimates of model parameters (we only include the coefficient matrix for simplicity as the ML estimation of innovation statistics are dependent on $\widehat{A_t}$):
\begin{equation}\label{eq:IC_general_form}
IC(p)=-\log(\mathcal{L}\left(\{\boldsymbol{X}_{t}^{(n)}\}_{t=1...T}|\widehat{A_t}\right))+\mathcal{P}(p)\,.
\end{equation}
The model order is selected as the order that minimizes the information criterion. 

For a stationary $d$-dimensional VAR(p) model with $N$ observations ($T=1$ reflecting the time-invariant model),
the penalty term $\mathcal{P}(p)$ involves the effect of model complexity by punishing on the number of parameters, which scales as $pd^2$. 
For AIC the penalty term is proportional to the number of parameters, scaling again as $\mathcal{P}(p)=pd^2$. 
BIC takes into account the effect of sample size $L$ as the $\log$-likelihood function also increases with the number of samples, and the corresponding penalty term is $\mathcal{P}(p)=\frac{1}{2} pd^2 \log(N)$.
As AIC often provides over-simplified models, we focus in the following sections on characterising BIC in different conditions.


\subsubsection{Derivation of the multi-trial ensemble BIC}
The challenge to apply BIC for the multi-trial case is to decide on the number of parameters and the sample size.
Naive ideas that there are $T\cdot pd^2$ parameters and $NT$ samples might be misleading as the multi-trial samples are modelled with different asymptotic assumptions.
Precisely, the stationary case assumes $N$ \textit{i.i.d.} samples from a time-invariant model with $pd^2$ parameters, while the time-varying model, consisting of $T \cdot pd^2$ parameters is estimated from $N$ \textit{i.i.d.} samples at each time point $t$ (with $t=1\dots T)$.
Despite previous attempts reviewed in Introduction, we propose here an extended version of BIC that is appropriate for non-stationary signals with multi-trial structures with the penalty term as $\mathcal{P}(p)=\frac{1}{2}T p d^2 \log(N)$).
The derivation is the following.

From a probabilistic perspective, the optimal model order $p$ should maximize the likelihood function given the data, i.e. maximize  $p(\{\boldsymbol{X}^{(n)}_t\}_{t=1...T}|p)$. 
This likelihood is different from the likelihood function we derived in Section~\ref{sec:var_estim}, i.e. $p(\{\boldsymbol{X}^{(n)}_t\}|\widehat{A_t},p)$ as the later is jointly parameterized by a point estimate $\widehat{A_t}$ of the $A_t$ matrices given the $p$-ordered model. 
By using instead a conditional  $p(\widehat{A_t}|p)$ for this set of parameters, we approximate the likelihood  $p(\{\boldsymbol{X}^{(n)}_t\}_{t=1...T}|p)$ with a Laplace approximation:
\begin{equation} \label{eq:Laplace approximation}
p(\{\boldsymbol{X}^{(n)}_t\}_{t=1...T}|p) 
= \prod_{t=1}^{T} \int p\left(\{\boldsymbol{X}^{(n)}_t\}, A_t|p\right)dA_t 
\approx \prod_{t=1}^{T}  p\left(\{\boldsymbol{X}^{(n)}_t\}|\widehat{A_t},p\right) p\left(\widehat{A_t}|p\right) (2\pi)^\frac{pd^2}{2} |H_t|^{-\frac{1}{2}} \,,
\end{equation}
with $H_t$ as the Hessian matrix of coefficient $A_t$ defined in Eq.~\ref{eq:hessian_compact_homo}.

Taking the logarithm of both sides, we can obtain the $\log$-likelihood as:
\begin{equation} \label{eq:logL_BIC}
\log p\left(\{\boldsymbol{X}^{(n)}_t\}_{t=1...T}|p\right) 
\approx \sum_{t=1}^{T}\log p\left(\{\boldsymbol{X}^{(n)}_t\}|\widehat{A_t},p\right) 
+\sum_{t=1}^{T} \log p\left(\widehat{A_t}|p\right) 
+ \frac{pd^2 T}{2}  \log (2\pi) 
-\frac{1}{2} \sum_{t=1}^{T} \log|H_t|\,.
\end{equation}

The likelihood function can be expressed using the ML-estimated parameters as:
\[
\sum_{t=1}^{T}\log p(\{\boldsymbol{X}^{(n)}_t\}|\widehat{A_t},p) 
= -\frac{ NTd}{2}\log(2 \pi) 
-\frac{N}{2}  \sum_{t=1}^{T}  \log|\widehat{\Sigma_t}| 
-\frac{1}{2} \sum_{t=1}^{T} \sum_{n=1}^{N} 
\left(\overline{\boldsymbol{X}}_{t}^{(n)}\!\!-\!\!\widehat{A_t} \overline{\boldsymbol{X}}_{p, t}^{(n)}\right)^{\top}
\widehat{\Sigma_t}^{-1} 
\left(\overline{\boldsymbol{X}}_{t}^{(n)}\!\!-\!\!\widehat{A_t} \overline{\boldsymbol{X}}_{p, t}^{(n)}\right)\,.
\]
The last term can be rewritten in the form of the sample-wise residual $\boldsymbol{\eta}_t^{(n)} = \overline{\boldsymbol{X}}_{t}^{(n)}-\widehat{A_t} \overline{\boldsymbol{X}}_{p, t}^{(n)}$ as
\[
-\frac{1}{2} \sum_{t=1}^{T} \sum_{n=1}^{N} 
\left(\overline{\boldsymbol{X}}_{t}^{(n)}-\widehat{A_t} \overline{\boldsymbol{X}}_{p, t}^{(n)}\right)^{\top}
\widehat{\Sigma_t}^{-1} 
\left(\overline{\boldsymbol{X}}_{t}^{(n)}-\widehat{A_t} \overline{\boldsymbol{X}}_{p, t}^{(n)}\right)
= -\frac{1}{2} \sum_{t=1}^{T} \sum_{n=1}^{N} \overline{\boldsymbol{\eta}}_t^{(n)\top} \widehat{\Sigma_t}^{-1} {\overline{\boldsymbol{\eta}}_t^{(n)}}.
\]
Asymptotically, $\mathbb{E}[\overline{\boldsymbol{\eta}}_t]=0$, $\mathbb{E}[\overline{\boldsymbol{\eta}}_t {\overline{\boldsymbol{\eta}}_t}^{\top}]=\widehat{\Sigma_t}$, thus this term converges to 
\begin{equation}
	-\frac{1}{2} \sum_{t=1}^{T} \sum_{n=1}^{N} \overline{\boldsymbol{\eta}}_t^{(n)\top} \widehat{\Sigma_t}^{-1} {\overline{\boldsymbol{\eta}}_t^{(n)}}\!\!\!
	= \!\!	-\frac{1}{2} NT \mathbb{E}[{\overline{\boldsymbol{\eta}}_t}^{\top} \widehat{\Sigma_t}^{-1} {\overline{\boldsymbol{\eta}}_t}]
	\!\!=\!\! -\frac{1}{2} NT (\Tr(\widehat{\Sigma_t}^{-1} \widehat{\Sigma_t})+\mathbb{E}[\overline{\boldsymbol{\eta}}_t]^{\top} \widehat{\Sigma_t}^{-1} \mathbb{E}[\overline{\boldsymbol{\eta}}_t])
\!\!	=\!\! -\frac{1}{2} NTd \,.
\end{equation}





Meanwhile, the determinant of the Hessian matrix is
\begin{equation}
|H_t| = |C_t|^d \cdot |\widehat{\Sigma_t}^{-1}|^{pd}\,.
\end{equation}

Then taking into account the assumptotic property of $C_t$ in Eq.~\ref{eq:assymp}, the logarithm of the determinant of the Hessian matrix is
\begin{equation}\label{eq:det_sum_Ht}
\log|H_t| = d \log|C_t| - pd\log|\widehat{\Sigma_t}|  = d \log | N \widehat{\Sigma}_{\boldsymbol{X}_{p}}| - pd \log|\widehat{\Sigma_t}|
=  pd^2 \log(N) + d \log|\widehat{\Sigma}_{\boldsymbol{X}_{p}}| -  p d \log|\widehat{\Sigma_t}| \,.
\end{equation}

Therefore, for the last term in Eq.~\ref{eq:det_sum_Ht}
\begin{equation}
\sum_{t=1}^{T}{ \log|H_t| } =  pd^2 T \log(N) + d \sum_{t=1}^{T}{\log|\widehat{\Sigma}_{\boldsymbol{X}_{p}}|} - pd (\sum_{t=1}^{T}{\log|\widehat{\Sigma_t}|})\,.
\end{equation}

As the trial number $N$ increases while trial length $T$ remains constant, the likelihood function in Eq.~\ref{eq:logL_BIC} can be approximated by the dominant changing terms:
\begin{equation} \label{eq:logL_BIC}
\log p(\{\boldsymbol{X}^{(n)}_t\}_{t=1...T}|p) 
\approx \sum_{t=1}^{T}\log p(\{\boldsymbol{X}^{(n)}_t\}|\widehat{A_t},p) 
-\frac{pd^2 T}{2}   \log(N)\,.
\end{equation}
 
Comparing with the general form of BIC in Eq.~\ref{eq:IC_general_form}, we now obtain a penalty term   $ pd^2 T\log(N)$, such that in our case the number of parameters of BIC is $ pd^2 T$ and the sample size is $N$. 
In practice, this can be estimated from data as 
\begin{equation} \label{eq:BIC_data}
\log p(\{\boldsymbol{X}^{(n)}_t\}_{t=1...T}|p) 
\approx -\frac{ NTd}{2}\log(2 \pi) 
	-\frac{N}{2}  \sum_{t=1}^{T}  \log|\widehat{\Sigma_t}| 
	-\frac{NTd}{2}  
	-\frac{pd^2 T}{2}  \log(N)\,.
\end{equation}
The quantity in Eq.~\ref{eq:BIC_data} defines what we refer to as "ensemble BIC" in later sections. A more precise way to determine the model order is to compare the complete form, which we refer to as "ensemble BIC (full)":
\begin{equation}\label{eq:BIC_full}
\log p(\{\boldsymbol{X}^{(n)}_t\}_{t=1...T}|p) 
\approx -\frac{ NTd}{2}\log(2 \pi) 
+\frac{pd-N}{2}  \sum_{t=1}^{T}  \log|\widehat{\Sigma_t}| 
-\frac{NTd}{2}  
-\frac{pd^2 T}{2}   \log(N) 
-\frac{d}{2}  \sum_{t=1}^{T}{\log|\widehat{\Sigma}_{\boldsymbol{X}_{p}}|}\,. 
\end{equation}
Normally, the ensemble BIC (but not the full version) is sufficient to obtain an optimized model order for empirical multi-trial panel data. 
We will also demonstrate in the Results section that the model order obtained by both ensemble BIC forms are almost identical.

\subsubsection{A special case for ensemble BIC within a 1-point time window}
Finally, we check the special case where the same multi-trial ensemble dataset is modelled by a time-invariant model VAR(p) model. In this case, Eq.~\ref{eq:logL_BIC} is revised as
\begin{equation} \label{eq:logL_BIC}
\log p(\{\boldsymbol{X}^{(n)}_t\}_{t=1...T}|p) 
\approx \log p(\{\boldsymbol{X}^{(n)}_t\}_{t=1...T}|\widehat{A_t},p) 
+ \log p(\widehat{A_t}|p) 
+ \frac{pd^2 }{2}  \log (2\pi) 
-\frac{1}{2}  \log|H_t|\,,
\end{equation}
where the last term is
\begin{equation}
\log|H| 
= d \log|N T \widehat{\Sigma}_{\boldsymbol{X}_{p}}| - pd\log|\widehat{\Sigma_t}|
= pd^2 \log(NT) + d \log| \widehat{\Sigma}_{\boldsymbol{X}_{p}}| - pd\log|\widehat{\Sigma_t}|\,.
\end{equation}

Therefore, the penalty term should be:  $pd^2 \log(NT)$, consistently with the classical BIC.

Theoretically there is not constraint to use the same model order for each 1-point time window, i.e. at each time point we are free to estimate an VAR(p) model with different model orders. However, finding a single optimal model order for the whole ensemble data facilitates the calculation of more advanced statistics, e.g. causality measures like the causal strength \citep{janzing_2013}. 


\section{Results}
In this section, we tested the proposed ensemble BIC for different types of panel data, ranging from simulational data with known model orders to experimental data. 
The ensemble BIC is well-validated with multi-trial ensembles where all trials are generated by the same time-varying model. 
We will also point out that the estimated model orders are overestimated when the trials are detected in the original signals and aligned by different variables of the VAR models.


\subsection{Validation on Perturbation Events based on VAR(4) models} \label{sec:perturb}
In this experiment, we simulated a bi-variate VAR(4) process $\boldsymbol{X}_t=[X_t^1, X^2_t]$ with uni-directional coupling ($X_t^2 \rightarrow X^2_t$). 
The autoregressive coefficient and innovation covariance matrices are designed to be constant across time, i.e. $ A_t =  A$, $\Sigma_t = \Sigma$:
\[
A = 
\begin{bmatrix} 
	-0.55 & 1.4 & -0.45 & -0.3 & -0.55 & 1.5 & -0.85  & 1.7 \\
	0 & 0.9 & 0 &-0.25 & 0 & 0 & 0 & 0.25 \\
\end{bmatrix}
, \Sigma=
\begin{bmatrix} 
	1 & 0 \\
	0 & 1 \\
\end{bmatrix}
\]
The events are constructed by designing a time-varying non-zero innovation mean as a deterministic perturbation of the system. Practically, at the peri-event window of the cause variable $X^2_t$ (centered at time ${t_n}$), the innovations mean is set as a 200-ms-long morlet-shaped waveform with the amplitude of 5 and 10, such that a strong oscillatory event is elicited in both variables. We refer to these events in the following as "Perturbation events".
We simulated 5000 such events in a long VAR(4) time series with an interval uniformly distributed between 200 and 800ms. 

\begin{figure}
\includegraphics[width=\textwidth]{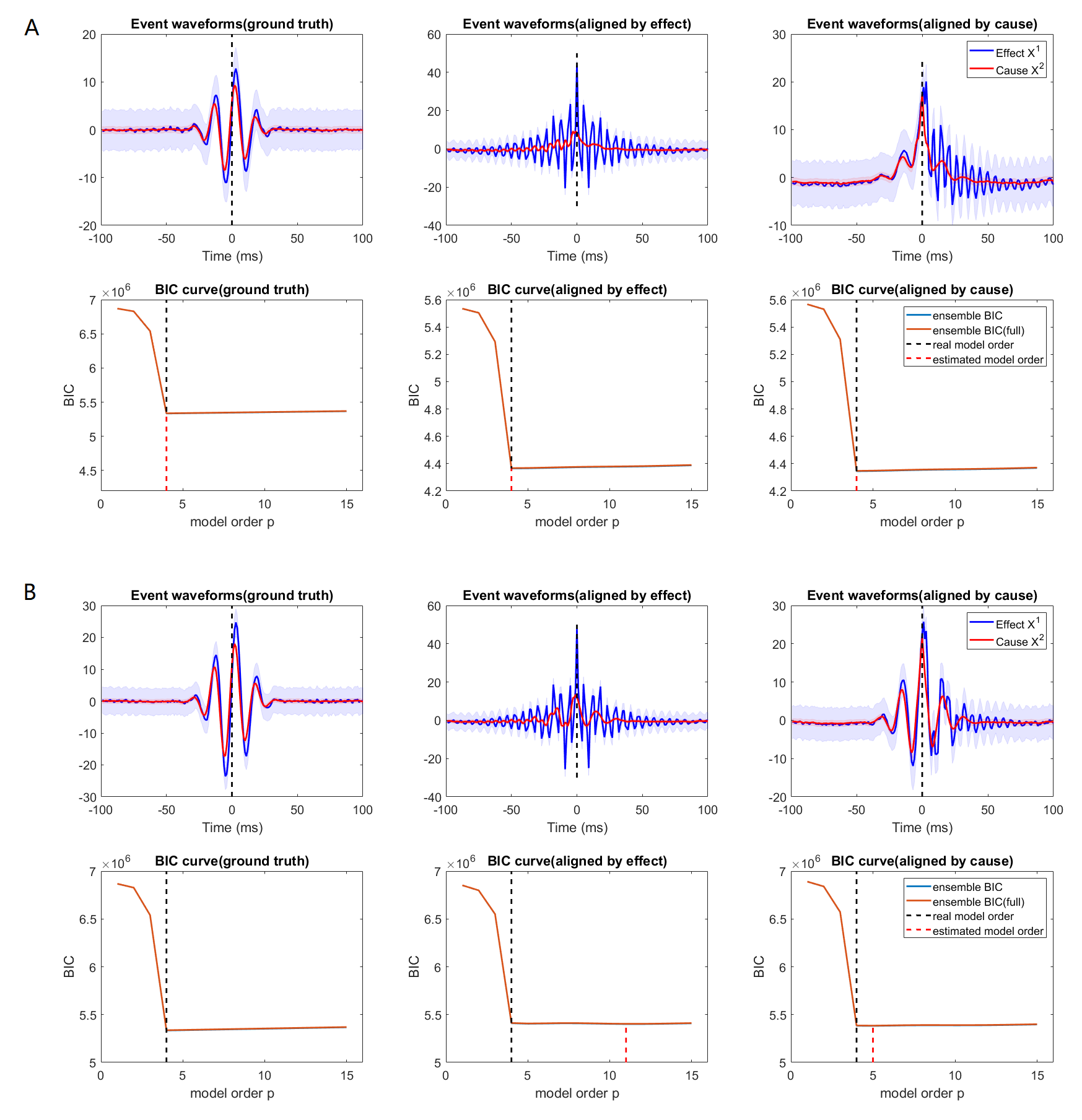}
\caption{\label{fig:perturb_events} 
	The ensemble BIC selects the correct model order for well-aligned panels, but may fail for panels aligned by sub-variables. 
	(A),(B) reflects the statistics of panel data and BIC results for different alignments with the amplitude of 5 (A) and 10 (B). 
	(top) Time-varying mean and variance of panels consisting the cause/effect variables aligned by ground truth reference points (left), aligned by the effect variable $X^1_t$ (middle) and aligned by the cause variable $X^2_t$.
	(bottom) Corresponding BIC curves as functions of model order $p$. Both the approximated BIC and the full version are plotted by blue and red curves. The real model order (4) is marked by black dashed lines, while the estimated model order are marked by red dashed lines.
	 }
\end{figure}

To validate the effectivity of the proposed ensemble BIC method for model order selection, we first extracted all the events based on the window surrouding their ground-truth reference points as designed, such that each trial is an indepedent sample of the same time-varying model. The statistics of the extracted ensemble are shown in Figure~\ref{fig:perturb_events}A, B (top left).
Figure~\ref{fig:perturb_events}A, B (bottom left) show the BIC curves for this ground-truth event panel, both for the proposed ensemble BIC as in Eq.~\ref{eq:logL_BIC} and Eq.~\ref{eq:BIC_data} and the full version as defined in Eq.~\ref{eq:BIC_full}. 
Clearly, the estimated model order coincides with the real model order 4, thus illustrating that the ensemble BIC is able to recover the real model order for multi-trial-panel data. 

However, in real data analysis of spontaneous events, one rarely has access to the real locations of the events. Therefore events are detected and aligned (a procedure as briefly described in the Methods section). 
Here we perform the BIC-based model order selection procedure on panel data detected and aligned based on different variables, i.e., either the cause variable $X^2_t$ and the effect variable $X^1_t$.

Based on either variable, the detection is performed by filtering the signal with a 49-ordered FIR filter in the range of 50-70Hz (as designed to be the feature frequency band of these perturbation events). 
A threshold of 3 times the standard deviation over the mean of the full filtered signal is then set to determine the rough location of the events, and the reference points ${t_n}$ are then determined by all time points over the threshold. Then the panel is collected by extracting the original signals in a peri-$t_n$ window of 200-ms length. The statistics of the panel aligned by the effect $X^1_t$ are shown in Figure~\ref{fig:perturb_events}A,B(top,middle) for perturbation amplitude of 5 and 10 separately, while corresponding statistics of the panel aligned by the cause $X^2_t$ is reflected in Figure~\ref{fig:perturb_events}A,B(top, right).

Interestingly, for the case where perturbation amplitude is 5, ensemble BIC selects the ground-truth model order fo all the three alignment conditions, while for the events where perturbation is stronger (amplitude=10), ensemble BIC fails to recover the real model orders.
Both estimated model orders for panels aligned by the effect and the cause are higher than the real model order, and the estimated model order for panels aligned by the effect is more overestimated than the other case. 
We will see that this phenomenon persists in following experiments. We will discuss the potential mechanisms in Discussions.

\subsection{Application on hippocample Sharp Wave-Ripples from CA3 to CA1}\label{sec:rodent_ripples}

Sharp Wave-Ripple (SPW-R) events, hypothesized as a key elements implementing memory consolidation in the brain, has been reported in the electrophysiological recordings within the hippocampus of both macaques and rodents. 

In this section, we applied the ensemble BIC method to an open source dataset where the CA3 and CA1 regions of rodent hippocampus has been recorded\cite{mizuseki2014neurosharing}. The coupling between the CA3 and CA1 regions has hypothetized to be uni-directional, i.e., CA3$\rightarrow$CA1, according to both anatomical basis \citep{CSICSVARI2000585} as well as data analysis results \citep{shajarisales2015telling}. Sharp Wave-Ripples (SPW-Rs) has been reported in both regions, while we detect SPW-Rs from the datasets and align by different regions, forming two panels for BIC-based model order selection.

The SPW-R events are detected from concatenated Local Field Potential (LFP) recordings in two sessions of a rat named 'vvp01' with the same electrode implantations. The sampling rate is 1252Hz. The signals are recorded with a 4-shank probe at each region consisting 8 channels for each shank, thus obtaining 32 channels for each region. 

\begin{figure}
	\includegraphics[width=\textwidth]{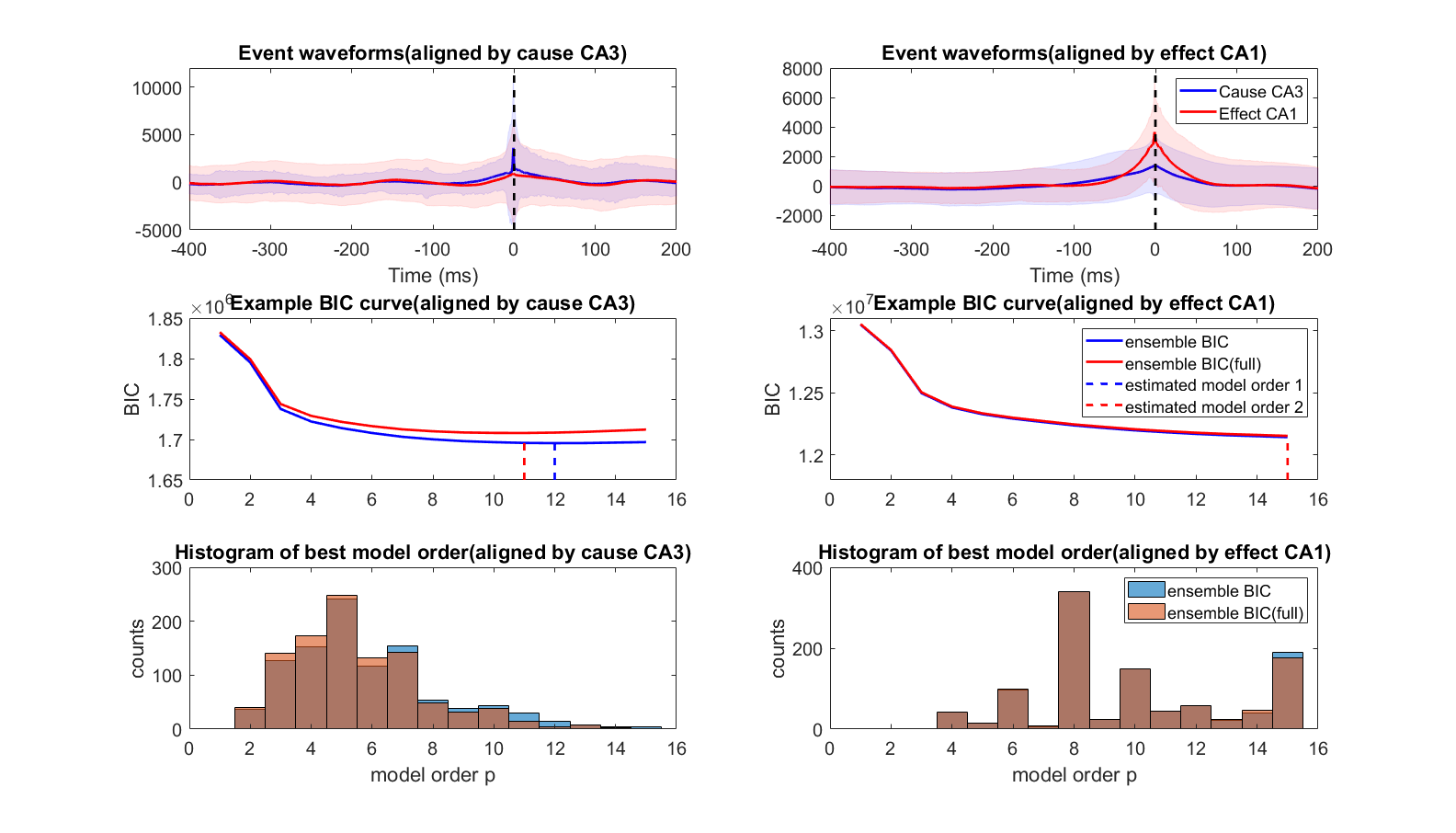}
	\caption{\label{fig:rodent_ripples} 
		The ensemble BIC applied to Sharp Wave-Ripples in rodent hippocampal CA3 and CA1 regions.
		(top) Time-varying mean and scaled variance of panels  aligned by the cause region CA3 (left) and aligned by effect region CA1 (right). 
		The variance is downscaled by 2000 to fit in the same figure with the amplitude of the mean.
		(middle) BIC curve as a funtion of model order $p$ for an example channle pair (no. 1024) for panels aligned by CA3 (left) and aligne by CA1 (right). Both the approximated BIC and the full version are plotted by blue and red curves. The best model orders obtained by ensemble BIC is marked by the blue dashed lines, while the best model orders obtained by the full version are marked by red dashed lines.
		(bottom) Histogram of best model orders obtained by the ensemble BIC and the full version for panels aligned by CA3 (left) and aligne by CA1 (right).
	}
\end{figure}

Following \cite{mizuseki2009theta}, the detection is performed with an 49-ordered FIR filter in the frequency band [140,230]Hz. Similar to Section~\ref{sec:perturb}, we set a threshold as 5 times the standard devidation over the mean of the filtered signals to locate the events and align them according to all points over the threshold. The peri-event window has been chosen to be [-400, 200]ms to fascilitate more advanced analysis after estimating the best model order. Panels for different alignment conditions are obtained by performing the detection procedure for a single-channel of either region and extracted with the combination of another channel in the other region. Therefore, we obtain 1024 panels for all channel pairs for each alignment conditions. The average statistics of all panels for each alignment condition are plotted in Figure~\ref{fig:rodent_ripples} (top).

Consistent with Section~\ref{sec:perturb}, for each alignment conditions, we estimated the optimized model order for each channel pair with the proposed ensemble BIC and the full version, resulting in 2 groups of 1024 best model orders. The BIC curves of an example channel pair with two alignment conditions are plotted in Figure~\ref{fig:rodent_ripples}(middle). The best orders estimated by the full version in Eq~\ref{eq:BIC_full} for the panel aligned by the cause region (CA3) is slightly smaller than the proposed ensemble BIC, although both are smaller than the best orders for panels aligned by the effect regions (CA1). The histograms displayed in Figure~\ref{fig:rodent_ripples}(bottom) match this observation, suggesting that the model order for panels aligned by the effect might be more biased. This is consistent with moder order rank in the perturbation events in Section~\ref{sec:perturb}.

\subsection{Application on hippocampal Sharp-Wave Ripples between the Striatum Radiatum and CA1}

Similar to Section~\ref{sec:rodent_ripples}, we applied the ensemble BIC methods to simultaneous recordings in macaque hippocampus in the region stratum radiatum and the pyramidal layers in CA1, where SPW-Rs can be detected by filtering the original signals in the frequency band 90-190Hz following a similar procedure.

In this dataset, 4 channels are identified for each region, thus leading to 16 channel pairs. We performed similar analysis procedure as in Similar to Section~\ref{sec:rodent_ripples}, resulting in 16 panels for each alignment condition. Thus similarly, the panel statistics averaged across channel pairs are represented in Figure~\ref{fig:monkey_ripples} (top). BIC curves in an example channel pairs for different alignment conditions are shown in Figure~\ref{fig:monkey_ripples} (middle), while the histograms of best model orders across channel pairs are plotted in Figure~\ref{fig:monkey_ripples} (bottom).

While the connection between the stratum radiatum and the CA1 pyramidal layers is considered to be uni-directional, previous studies (\cite{RAMIREZVILLEGAS20181224}) suggested with modelling approaches the existence of interactions in the opposite direction. Regardless of the true causal direction, we still observe slightly higher best model orders when the panels are aligned by the pyramidal layer compared to aligned by the stratum radiatum.

\begin{figure}
	\includegraphics[width=\textwidth]{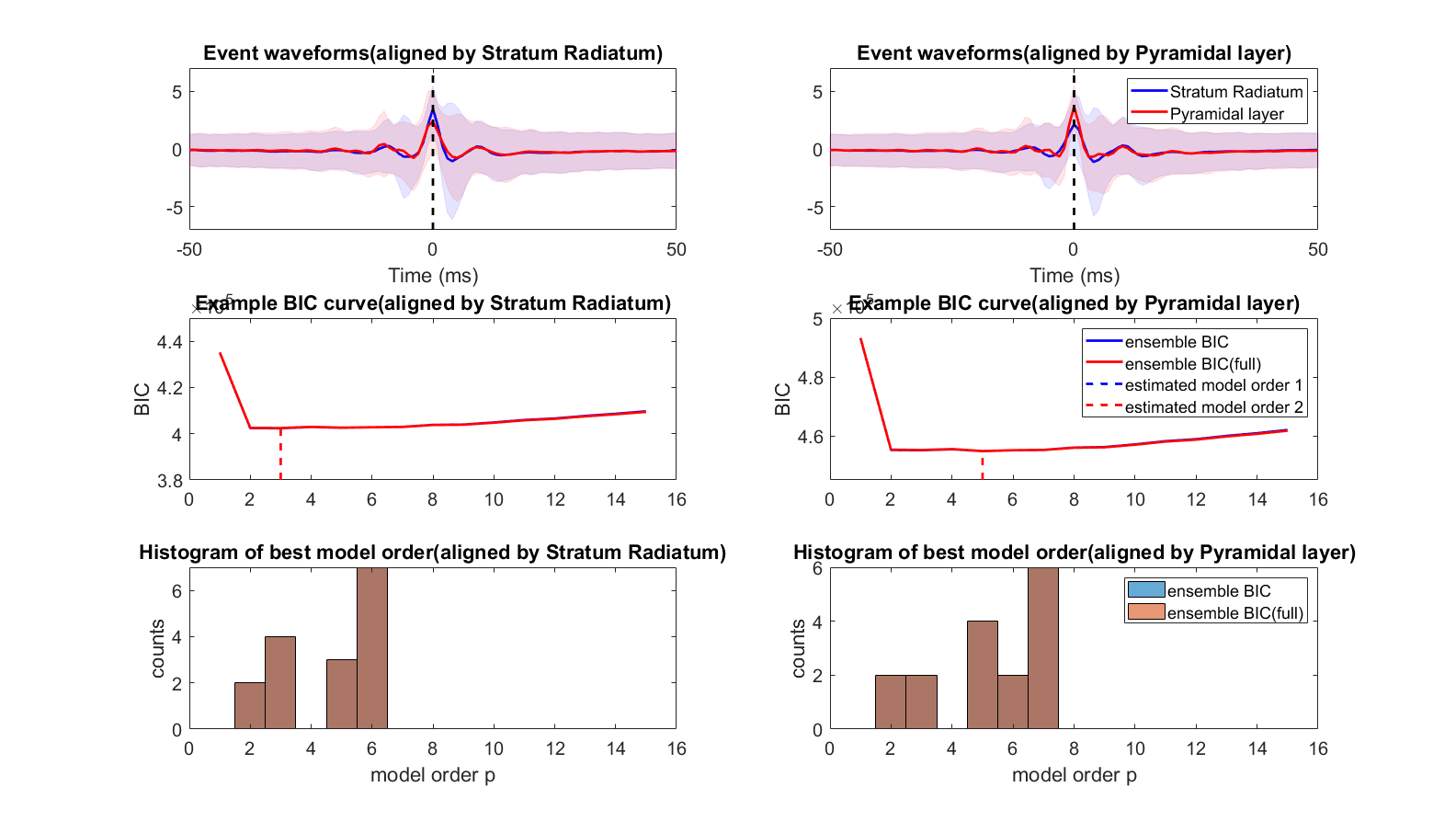}
	\caption{\label{fig:monkey_ripples} 
		The ensemble BIC applied to Sharp Wave-Ripples in two regions of monkey hippocampus: the stratum radiatum and pyramidal layer of CA1.
		(top) Time-varying mean and scaled variance of panels  aligned by the stratum radiatum (left) and aligned by pyramidal layer (right). 
		(middle) BIC curve as a funtion of model order $p$ for an example channle pair (no. 6) for panels aligned by stratum radiatum (left) and aligne by pyramidal layer (right). Both the approximated BIC and the full version are plotted by blue and red curves. The best model orders obtained by ensemble BIC is marked by the blue dashed lines, while the best model orders obtained by the full version are marked by red dashed lines.
		(bottom) Histogram of best model orders obtained by the ensemble BIC and the full version for panels  aligned by stratum radiatum (left) and aligne by pyramidal layer (right).
	}
\end{figure}

\section{Discussion}

In general, we have adapted the well-documented BIC method for model order selection to multi-trial ensemble data. based on VAR(p) models. Specifically, we propose that the equivalent parameters in the classical time-invariance BIC form to be $pd^2T$ for the number of parameters and $N$ to be the sample size for time-varying panel models. The BIC methods has been validated with multi-trial events simulated with an time-varying VAR(4) models, suggesting that the proposed ensemble BIC is an effective tool for determining complexity in dynamical modelling.

\begin{figure} \label{fig:bias}
	\includegraphics[width=\textwidth]{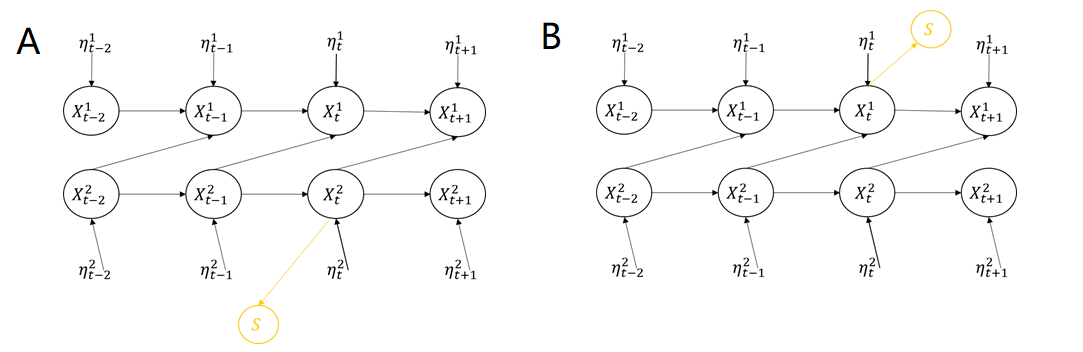}
	\caption{\label{fig:monkey_ripples} 
		Illustration of the effect of different alignment conditions.
		(A) A causal diagram for an example VAR(1) model with uni-directional coupling. The yellow node S, dependent on the cause variable, represents a selection node for detecting the events and choose the data points over a certain threshold.
		(B) The same causal diagram as in (A) but with a selection node that depends on the effect variable.
	}
\end{figure}

Besides, we found that when different trials in a panel is aligned by one of the variables, the ensemble BIC fail to recover the true model order. However, this is critical as aligning is a common precedure for analyzing spontaneous events, thus should be addressed to achieve a practical use of the ensemble BIC. Here we briefly discuss the underlying mechanism. From the perspective of a causal diagram, as seen in Figure~\ref{fig:bias}, alignment is equivalent to adding a selection node to either the cause or the effect variable. After detection with a thresholding procedure, the S node can be seen as observed, thus affecting the other dependencies in the diagram (reflected by the errors). Simply put, with d-separation it is clear that observing S induces spurious dependencies between nodes before time t but does not affect the nodes after t. Therefore, the VAR model estimation will be biased, leading to wrong estimation of the likelihood function in e.g., Eq.~\ref{eq:BIC_data}. Aligning by the cause is slightly less biased compared to aligning by the effect because the coupling dependence is not affected by the selection. 
Therefore, further studies should be conducted to correct for the selection bias in order to obtain an appropriate model order to model the time-varying dynanmics in modelling event-based panels.

\bibliographystyle{unsrtnat}
\bibliography{snapshot}  






\end{document}